\documentclass{article}

\usepackage[preprint]{neurips_2022}

\usepackage[utf8]{inputenc}
\usepackage[T1]{fontenc}

\usepackage{newtxtext,newtxmath}

\usepackage[dvipsnames]{xcolor}   
\definecolor{linkColor}{rgb}{0.2,0.4,0.6} 
\usepackage{graphicx}
\usepackage{booktabs}
\usepackage{amsmath}
\usepackage{mathtools}
\usepackage{nicefrac}
\usepackage{multirow}
\usepackage{wrapfig}
\usepackage{float}
\usepackage{bbm}
\usepackage{enumitem}
\usepackage{url}
\usepackage{pifont}
\usepackage{wasysym}
\usepackage{arydshln}
\usepackage{listings}
\usepackage[most]{tcolorbox}      
\usepackage{adjustbox}
\usepackage{setspace}
\usepackage[normalem]{ulem}       
\usepackage{capt-of}
\usepackage{algorithm}
\usepackage{algorithmic}          
\usepackage{mdframed}
\usepackage{subcaption}           
\usepackage{makecell}
\usepackage{colortbl}
\usepackage{hyperref}

\hypersetup{
  colorlinks=true,
  linkcolor=linkColor,
  citecolor=linkColor,
  urlcolor=linkColor
}

\usepackage{amssymb}

\usepackage{amsthm}
\usepackage{siunitx}

\definecolor{softred}{RGB}{214, 96, 96}
\definecolor{softblue}{RGB}{93, 133, 195}
\definecolor{softteal}{RGB}{72, 168, 154}
\definecolor{sgray}{gray}{0.95}

\newcommand{\showgap}[2]{\textcolor{#2}{\scriptsize{(+#1)}}}

\usepackage{tikz}
\usepackage{pgfplots}
\pgfplotsset{compat=1.18}

\newcommand{\ICMLTableCaption}[1]{%
  \setlength{\abovecaptionskip}{0.1in}%
  \setlength{\belowcaptionskip}{0.1in}%
  \caption{\fontsize{9}{11}\selectfont #1}%
}

\usepackage[capitalize,noabbrev]{cleveref}

\theoremstyle{plain}

\theoremstyle{definition}

\theoremstyle{remark}

\lstdefinestyle{mypython}{
  language=Python,
  basicstyle=\ttfamily\footnotesize,
  keywordstyle=\bfseries,
  commentstyle=\itshape,
  stringstyle=,
  columns=fullflexible,
  keepspaces=true,
  showstringspaces=false,
  tabsize=4,
  breaklines=true,
}

\title{Sparse-BitNet: 1.58-bit LLMs are Naturally Friendly\\ to Semi-Structured Sparsity}

\author{%
\textbf{Di Zhang}\textsuperscript{1,2}\quad
\textbf{Xun Wu}\textsuperscript{1}\quad
\textbf{Shaohan Huang}\textsuperscript{1}\quad
\textbf{Yudong Wang}\textsuperscript{2}\quad
\textbf{Hanyong Shao}\textsuperscript{1,2}\quad
\textbf{Yingbo Hao}\textsuperscript{1,3}\quad\\
\textbf{Zewen Chi}\textsuperscript{1}\quad
\textbf{Li Dong}\textsuperscript{1}\quad
\textbf{Ting Song}\textsuperscript{1}\quad
\textbf{Yan Xia}\textsuperscript{1}\quad
\textbf{Zhifang Sui}\textsuperscript{2}\quad
\textbf{Furu Wei}\textsuperscript{1}\\[0.4em]
\textsuperscript{1}\,Microsoft Research \quad
\textsuperscript{2}\,Peking University \quad
\textsuperscript{3}\,South China University of Technology\\
\href{https://aka.ms/GeneralAI}{https://aka.ms/GeneralAI}
}

\begin{document}
\maketitle

\begin{abstract}
Semi-structured N:M sparsity and low-bit quantization (e.g., 1.58-bit BitNet) are two promising approaches for improving the efficiency of large language models (LLMs), yet they have largely been studied in isolation. In this work, we investigate their interaction and show that \textbf{1.58-bit BitNet is naturally more compatible with N:M sparsity than full-precision models}.
To study this effect, we propose \emph{Sparse-BitNet}, a unified framework that jointly applies 1.58-bit quantization and dynamic N:M sparsification while ensuring stable training for the first time. 
Across multiple model scales and training regimes (sparse pretraining and dense-to-sparse schedules), 1.58-bit BitNet consistently exhibits smaller performance degradation than full-precision baselines at the same sparsity levels and can tolerate higher structured sparsity before accuracy collapse.
Moreover, using our custom sparse tensor core, Sparse-BitNet achieves substantial speedups in both training and inference, reaching up to 1.30$\times$. These results highlight that combining extremely low-bit quantization with semi-structured N:M sparsity is a promising direction for efficient LLMs. Code available at \href{https://github.com/AAzdi/Sparse-BitNet}{https://github.com/AAzdi/Sparse-BitNet}
\end{abstract}

\section{Introduction}


Large Language Models (LLMs) have demonstrated remarkable performance across a wide range of tasks~\citep{gpt4,gpt5,deepseekv3,qwen3}. However, their rapidly increasing scale leads to substantial training and inference costs~\citep{deepseekv3,qwen3}, making efficiency a central challenge in modern LLMs research. Among existing approaches, \emph{Quantization}~\citep{lin2024awq,frantar2022gptq,dettmers2022int8} and \emph{Sparsity}~\citep{jaszczur2021sparse,torchao,hu2024accelerating} have emerged as two widely studied and effective strategies for improving LLM efficiency.

Semi-structured N:M sparsity, particularly the widely supported 2:4 pattern~\citep{fang2024maskllm,hu2024accelerating,torchao}, has attracted increasing attention for its ability to accelerate matrix multiplication on NVIDIA Sparse Tensor Cores~\citep{nvidia_hopper_whitepaper}, which require that at most 2 out of every 4 weights are non-zero to exploit sparsity for hardware acceleration. Nevertheless, existing works~\citep{fang2024maskllm,haziza2025accelerating,kubler2025proximal} have primarily applied semi-structured sparsity to full-precision LLMs. Under strict N:M constraints, these full-precision models often suffer rapid accuracy degradation, making it challenging to achieve both high sparsity and high performance simultaneously.

\begin{figure}[t]
    \centering
    \includegraphics[width=\linewidth]{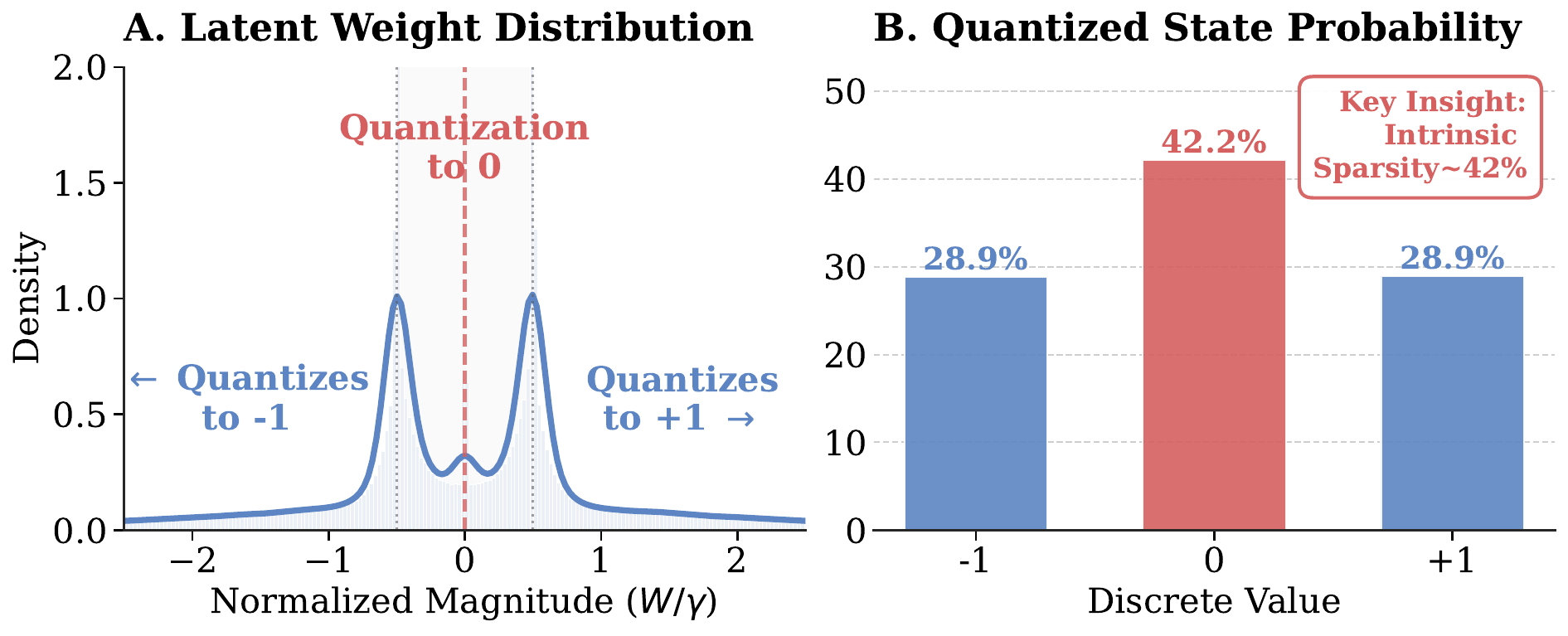}
    \caption{\textbf{Intrinsic Sparsity in 1.58-bit BitNet.} 
    We present the aggregated weight statistics averaged across all linear layers of the pre-trained 1.58-bit BitNet (2B) model~\citep{ma2025bitnetb158}.
    {(A)} The distribution of normalized latent weights exhibits a distinct quantization-valley structure, where the majority of values fall within the [-0.5, 0.5] rounding interval.
    {(B)} Consequently, \textbf{the quantized discrete states are dominated by zeros (approx. 42.3\%), confirming that BitNet naturally converges to a highly sparse representation without explicit pruning}.}
    \label{fig:intro_sparsity}
    \label{fig:intro_bitnet_2B4T}
\end{figure}

\begin{figure}[t]
    \centering
    \includegraphics[width=\linewidth]{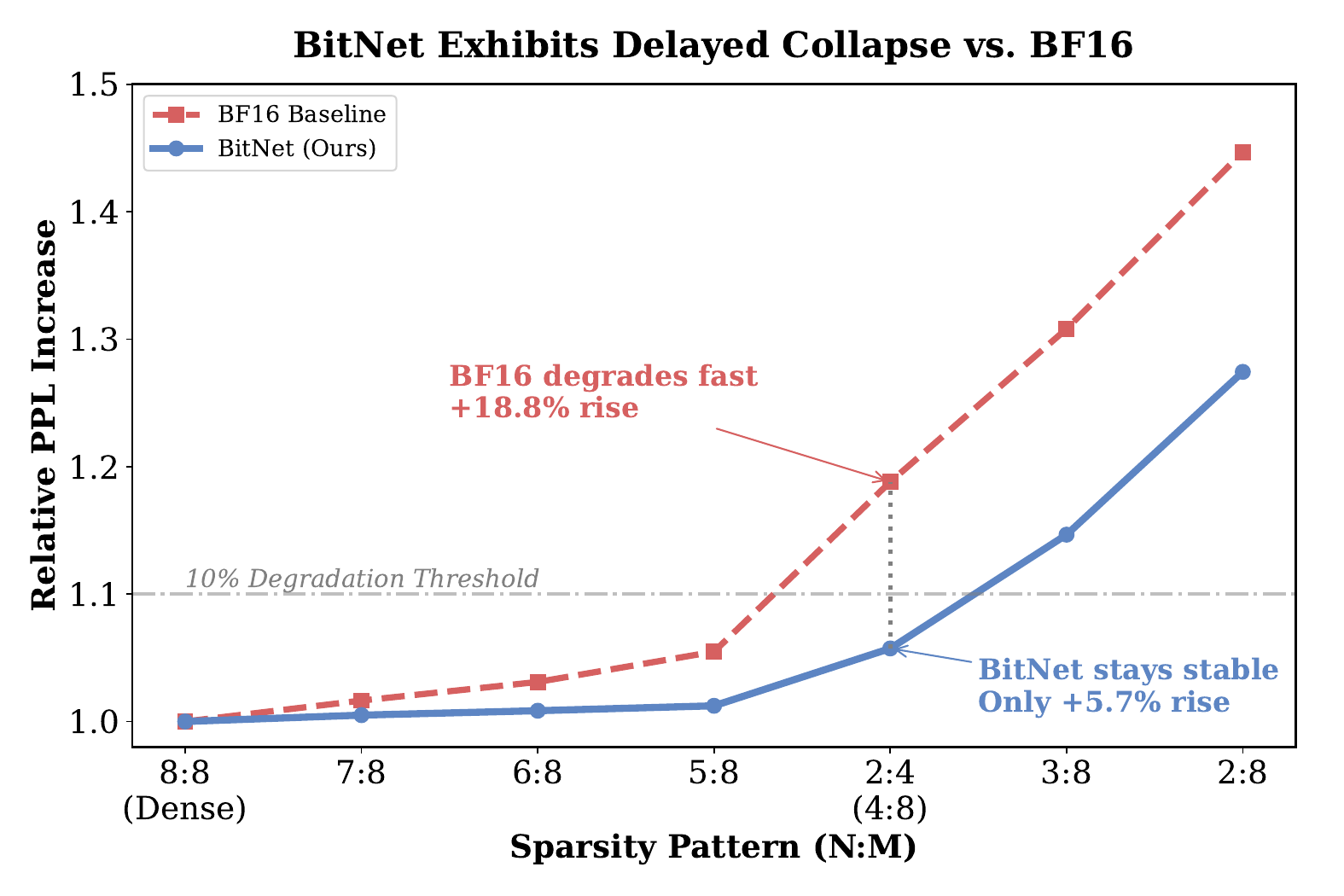}
    \caption{\textbf{Comparison of N:M sparsity friendliness between 1.58-bit BitNet and full-precision models.}
    Normalized PPL increase relative to each method's dense (8:8) counterpart.
    The dashed line marks a 10\% degradation threshold.
    At 2:4 (50\% sparsity; same ratio as 4:8), BF16 exceeds the threshold (+18.8\%) while BitNet remains below it (+5.7\%), indicating 1.58-bit BitNet is  more sparsity-friendly than full-precision models.}
    \label{fig:ppl_sparsity_trend}
\end{figure}

In parallel, extremely low-bit quantization has emerged as an alternative pathway toward LLM efficiency. In particular, the 1.58-bit BitNet~\citep{wang2023bitnet,wang2025bitnet,ma2025bitnet} quantizes weights into a ternary set $\{-1, 0, 1\}$ and achieves performance competitive with full-precision baselines at scale~\citep{ma2025bitnet}. 
As shown in Figure~\ref{fig:intro_sparsity}, the weights of a pretrained 1.58-bit BitNet exhibit a distinctive quantization-valley structure, with a high fraction of zero-valued ternary weights (approximately 42\%), naturally forming a sparse representation without explicit pruning. Although these zeros are unstructured and do not directly enable N:M sparse kernel acceleration, they reveal a weight-magnitude geometry that is inherently more compatible with magnitude-based N:M sparsity selection.

This observation highlights an important yet underexplored gap in the literature. While prior work has extensively studied N:M sparsity in full-precision models and low-bit quantization in isolation, the interaction between extremely low-bit quantization and semi-structured sparsity remains largely unexplored. 
This gap motivates the following research question:

\begin{quote}
\textbf{\emph{Under the same N:M sparsity constraints, is 1.58-bit BitNet more sparsity-friendly than full-precision models?}}
\end{quote}



To address this question, we first develop a unified Sparse-BitNet training framework (\S~\ref{sec:method}) that jointly enforces 1.58-bit BitNet weight quantization and N:M semi-structured sparsity constraints while ensuring stable training of LLMs.


Using this framework, we compare the \emph{sparsity-friendliness} of 1.58-bit BitNet and BF16 models under two settings:
(1) from-scratch pretraining with N:M sparsity constraints, and
(2) dense-to-sparse training schedules that switch from dense to N:M sparse training at different stages. Experiments on the Qwen-2.5 model family~\citep{qwen2.5} across multiple scales (0.5B–3B) show that 1.58-bit BitNet consistently incurs smaller performance degradation than BF16 under identical sparsity constraints (see in Figure~\ref{fig:ppl_sparsity_trend}).

In addition, we measure end-to-end inference throughput on NVIDIA GPUs using our in-house 6:8 sparse operator across varying sequence lengths and batch sizes, and find that combining 1.58-bit BitNet with N:M sparsity yields substantial speedups (reaching up to 1.30$\times$) in both training and inference. Our main contributions are summarized as follows:

\begin{itemize}
  \item We first systematically investigate and show that extremely low-bit quantization (e.g., 1.58-bit BitNet) is inherently more compatible with semi-structured N:M sparsity than full-precision (e.g., BF16) models, exhibiting smaller accuracy degradation under identical sparsity constraints.
  \item We propose a Sparse-BitNet training framework that jointly integrates 1.58-bit quantization and N:M sparsity to improves training stability and robustness (\S~\ref{sec:method}).
  \item Extensive experiments demonstrate that Sparse-BitNet achieves better accuracy–efficiency trade-offs than sparse full-precision baselines (see in Figure~\ref{fig:ppl_sparsity_trend}), with favorable end-to-end speedups (\S~\ref{sec:experiments}).
\end{itemize}

\section{Sparse-BitNet}
\label{sec:method}

In this section, we present Sparse-BitNet, a method that integrates semi-structured sparsity into the ternary quantization training landscape. We first review the foundational components---BitNet b1.58 and $N\!:\!M$ sparsity---and then detail our proposed architecture, training strategy, and the interaction between these two compression dimensions.

\subsection{Preliminaries}
\label{sec:method:preliminaries}

\noindent\textbf{1.58-bit BitNet.}
1.58-bit BitNet~\cite{ma2025bitnetb158} is an evolution of the BitNet architecture~\cite{wang2023bitnet} that constrains weights to a ternary set $\mathcal{T} = \{-1, 0, +1\}$, resulting in a theoretical information density of $\log_2 3 \approx 1.58$ bits per parameter. 
The fundamental building block is the \emph{BitLinear} layer. 
For a weight matrix $\mathbf{W} \in \mathbb{R}^{d_{\text{out}} \times d_{\text{in}}}$, the ternary weights $\mathbf{W}_q$ are derived by scaling the latent full-precision weights by their average absolute magnitude:
\begin{equation}
\mathbf{W}_q = \mathrm{RoundClip}\left(\frac{\mathbf{W}}{\gamma + \epsilon}, -1, +1\right),
\end{equation}
where $\gamma = \frac{1}{d_{\text{out}}d_{\text{in}}}\|\mathbf{W}\|_1$ serves as the scaling factor and $\epsilon$ is a small constant for numerical stability. 
Activations are quantized to 8-bit integers using absmax scaling: $\tilde{\mathbf{x}} = \mathrm{Quant}(\mathbf{x}) = \mathrm{Clip}(\mathbf{x} \cdot \frac{127}{\max(|\mathbf{x}|) + \epsilon}, -128, 127)$. 
The forward pass is then computed as $\mathbf{y} \approx \gamma \cdot \mathrm{matmul}(\mathbf{W}_q, \tilde{\mathbf{x}})$, significantly reducing the computational overhead by replacing floating-point multiplications with integer additions and scaling.

\noindent\textbf{Semi-Structured N:M Sparsity.}
Semi-structured sparsity enforces a fine-grained pattern where at most $N$ elements are non-zero out of every $M$ consecutive weights~\cite{zhou2021nm}.
This format retains hardware efficiency (e.g., via Sparse Tensor Cores) while offering flexibility over coarse pruning.
For a weight matrix $\mathbf{W}$, we define a binary mask $\mathbf{M} \in \{0, 1\}^{d_{\text{out}} \times d_{\text{in}}}$ such that for every group of $M$ elements (typically along the input dimension), $\|\mathbf{M}_{\text{group}}\|_0 = N$.
The sparse computation is performed as $\mathbf{y} = (\mathbf{W} \odot \mathbf{M})\mathbf{x}$.
In this work, we focus on the $6\!:\!8$ pattern ($25\%$ sparsity) as a balanced trade-off between compression and accuracy for low-bit LLMs, while also benchmarking against the standard $2\!:\!4$ pattern.

\subsection{The Sparse-BitLinear Architecture}
\label{sec:method:sparse_bitlinear}

Sparse-BitNet replaces standard linear projections with the \emph{Sparse-BitLinear} layer. This layer composes ternary quantization and $N\!:\!M$ masking into a single operator trained from scratch. We maintain a high-precision master weight $\mathbf{W}$ (e.g., in \texttt{bf16}) during optimization to accumulate gradients.

\noindent\textbf{Magnitude-based Mask Generation.}
To determine the sparsity pattern, we compute the mask $\mathbf{M}_{N:M}$ directly from the master weights $\mathbf{W}$. We employ magnitude pruning: for every contiguous group of size $M$, we select the indices of the $N$ largest absolute values.
\begin{equation}
\mathbf{M}_{N:M} = \Pi_{N:M}\bigl(|\mathbf{W}|\bigr),
\label{eq:mask_generation}
\end{equation}
where $\Pi_{N:M}(\cdot)$ is the per-group Top-$N$ indicator function. Crucially, masking is performed based on the \emph{pre-quantized} weights to preserve fine-grained magnitude rankings, avoiding the tie-breaking issues that would arise if masking were applied to discrete ternary values.

\noindent\textbf{Quant-and-Mask Computation.}
The forward pass follows a ``quant-and-mask'' paradigm.
We first quantize the activations $\mathbf{x}$ to $\tilde{\mathbf{x}}$ (typically 8-bit) and the master weights $\mathbf{W}$ to ternary values $\mathbf{W}_q = Q_t(\mathbf{W}) \in \{-1, 0, +1\}$.
We then apply the mask to the quantized weights:
\begin{equation}
\mathbf{W}_{\mathrm{eff}} = \mathbf{W}_q \odot \mathbf{M}_{N:M}.
\label{eq:effective_weights}
\end{equation}
The output is computed using these effective sparse-quantized weights:
\begin{equation}
\mathbf{y} \approx s \cdot \bigl(\mathbf{W}_{\mathrm{eff}}\,\tilde{\mathbf{x}}\bigr),
\label{eq:forward_pass}
\end{equation}
where $s$ absorbs the quantization scales. This order ensures that the N:M mask pattern is enforced on the final discrete weights used for inference, yielding well-defined N:M metadata/layout for hardware kernels. We provide detailed torch-style pseudo code in Algorithm~\ref{alg: sparse_bitlinear} for reference.

\subsection{Training Strategy}
\label{sec:method:training}

We train Sparse-BitNet from scratch, optimizing the master weights $\mathbf{W}$ end-to-end. The training procedure is summarized in Algorithm~\ref{alg:sparse_bitnet_train}.

\noindent\textbf{Dynamic Mask Recomputation.}
Unlike post-training pruning methods, we recompute $\mathbf{M}_{N:M}$ at \emph{every training step} using Eq.~\eqref{eq:mask_generation}. This corresponds to a projected optimization approach where the discrete constraint is re-evaluated continuously, preventing mask staleness and allowing the topology of the network to evolve alongside the weight values.

\noindent\textbf{Gradient Estimation via Dual STE.}
Since both the quantization function $Q_t(\cdot)$ and the mask selection $\Pi_{N:M}(\cdot)$ are non-differentiable, we require an approximation to propagate gradients to the master weights $\mathbf{W}$.
We employ a \textbf{Dual Straight-Through Estimator (STE)} approach.
First, for the ternary quantizer $Q_t$, we pass gradients through as the identity function within the clipping range, following standard BitNet training.
Second, and crucially, we also apply STE to the sparsity mask.
Specifically, we treat the mask operator as transparent during the backward pass:
\begin{equation}
\frac{\partial \mathcal{L}}{\partial \mathbf{W}}
\;=\;
\frac{\partial \mathcal{L}}{\partial \mathbf{W}_{\mathrm{eff}}}
\cdot
\frac{\partial \mathbf{W}_{\mathrm{eff}}}{\partial \mathbf{W}}
\;\approx\;
\frac{\partial \mathcal{L}}{\partial \mathbf{W}_{\mathrm{eff}}},
\label{eq:ste_grad}
\end{equation}

This means that gradients flow to \emph{all} master weights, including those that were pruned (masked out) in the forward pass.
This differs from methods that gate gradients with the mask (i.e., $\frac{\partial \mathcal{L}}{\partial \mathbf{W}} \approx \mathbf{M} \odot \frac{\partial \mathcal{L}}{\partial \mathbf{W}_{\mathrm{eff}}}$)~\cite{zhou2021nm}, which restricts updates only to the active set.
By allowing \emph{dense gradient updates}, our method enables pruned weights to receive direct feedback and potentially grow large enough to re-enter the Top-$N$ set in subsequent steps, preventing premature structural collapse.
We empirically validate the necessity of this dense gradient flow in our ablation studies (see in \S~\ref{sec:gradient}).

\begin{algorithm}[t]
\caption{Training Sparse-BitNet (Ternary, Per-step Mask, Dual STE)}
\label{alg:sparse_bitnet_train}
\begin{algorithmic}[1] 
    \STATE {\bfseries Require:} Data $\mathcal{D}$, master weights $\mathbf{W}$, pattern $(N,M)$
    \FOR{step $t = 1, 2, \dots$}
        \STATE $\mathcal{B} \leftarrow \text{Sample}(\mathcal{D})$
        \STATE $\mathbf{M}_{N:M} \leftarrow \Pi_{N:M}(|\mathbf{W}|)$ \quad {\color{gray!40}//Compute Mask}
        \STATE $\tilde{\mathbf{x}} \leftarrow Q_a(\mathrm{Norm}(\mathbf{x}))$
        \STATE $\mathbf{W}_q \leftarrow Q_t(\mathbf{W})$
        \STATE $\mathbf{W}_{\mathrm{eff}} \leftarrow \mathbf{W}_q \odot \mathbf{M}_{N:M}$ \quad {\color{gray!40}// Apply Mask}
        \STATE $\mathbf{y} \approx s \cdot \mathbf{W}_{\mathrm{eff}}\,\tilde{\mathbf{x}}$ \quad {\color{gray!40}// Forward Pass}
        \STATE \textbf{Backward:}
        \STATE \quad (1) STE through $Q_t(\cdot)$
        \STATE \quad (2) STE through $\Pi_{N:M}(\cdot)$ (Do \textbf{not} mask gradients)
        \STATE Update \textbf{all} master weights $\mathbf{W}$ via optimizer
    \ENDFOR
\end{algorithmic}
\end{algorithm}

\section{Experiments}
\label{sec:experiments}

\begin{table*}[t]
\centering
\renewcommand{\arraystretch}{1.15} 
\setlength{\tabcolsep}{8pt}        

\caption{\textbf{Downstream task performance evaluation.} We compare the performance of Dense and Sparse models across varying scales (0.5B, 1.5B, and 3B) on five benchmarks. The $\Delta$ column highlights the performance drop post-sparsification. \textbf{Notable Trend: BitNet consistently demonstrates superior resilience to sparsification (smaller $\Delta$ drop) compared to BF16 across all model scales}.}
\label{tab:downstream_delta_soft}

\small
\begin{tabular}{l ccccc c cc}
\toprule
\multirow{2}{*}{\textbf{Method}} & \multicolumn{5}{c}{\textbf{Task Accuracy (\%)}} & & \multicolumn{2}{c}{\textbf{Average}} \\
\cmidrule(lr){2-6} \cmidrule(lr){8-9}
 & HellaSwag & ARC-E & PIQA & BoolQ & COPA & & Score & \textbf{$\Delta$} \\
\midrule

\rowcolor{sgray} \multicolumn{9}{l}{\textit{\textbf{Qwen2.5-0.5B}}} \\
\hspace{1em}Dense BF16              & 40.45 & 43.31 & 69.04 & 60.12 & 71.00 & & 56.78 & -- \\
\hspace{1em}Sparse BF16 (6:8)       & 39.21 & 39.84 & 66.43 & 57.32 & 66.00 & & 53.76 & \textcolor{softred}{-3.02} \\
\addlinespace[0.2em]
\hspace{1em}Dense BitNet            & 35.27 & 40.70 & 65.07 & 59.24 & 69.00 & & 53.86 & -- \\
\hspace{1em}Sparse BitNet (6:8)     & 34.95 & 37.63 & 63.87 & 59.08 & 68.00 & & 52.71 & \textbf{\textcolor{softblue}{-1.15}} \\
\addlinespace[0.5em]

\rowcolor{sgray} \multicolumn{9}{l}{\textit{\textbf{Qwen2.5-1.5B}}} \\
\hspace{1em}Dense BF16              & 49.32 & 48.65 & 72.47 & 60.28 & 71.00 & & 60.34 & -- \\
\hspace{1em}Sparse BF16 (6:8)       & 40.44 & 39.73 & 67.19 & 47.77 & 68.00 & & 52.63 & \textcolor{softred}{-7.71} \\
\addlinespace[0.2em]
\hspace{1em}Dense BitNet            & 44.64 & 44.61 & 70.29 & 57.43 & 70.00 & & 57.39 & -- \\
\hspace{1em}Sparse BitNet (6:8)     & 36.95 & 40.57 & 65.23 & 55.26 & 70.00 & & 53.60 & \textbf{\textcolor{softblue}{-3.79}} \\
\addlinespace[0.5em]

\rowcolor{sgray} \multicolumn{9}{l}{\textit{\textbf{Qwen2.5-3B}}} \\
\hspace{1em}Dense BF16              & 54.88 & 51.77 & 73.67 & 61.56 & 75.00 & & 63.38 & -- \\
\hspace{1em}Sparse BF16 (6:8)       & 52.87 & 50.52 & 71.58 & 53.91 & 72.00 & & 60.18 & \textcolor{softred}{-3.20} \\
\addlinespace[0.2em]
\hspace{1em}Dense BitNet            & 50.46 & 48.23 & 71.93 & 53.18 & 70.00 & & 58.76 & -- \\
\hspace{1em}Sparse BitNet (6:8)     & 51.20 & 47.32 & 71.93 & 51.35 & 68.00 & & 57.96 & \textbf{\textcolor{softblue}{-0.80}} \\
\bottomrule
\end{tabular}
\end{table*}

\subsection{Experimental Setup}
\label{sec:exp_setup}

\noindent\textbf{Backbone models.}
We study three model scales based on the Qwen2.5~\citep{qwen2.5} architecture: Qwen2.5-0.5B, Qwen2.5-1.5B, and Qwen2.5-3B. Unless otherwise noted, all models are trained from scratch under the same data mixture and token budget.

\noindent\textbf{Sparsity pattern.}
Our main sparse-training experiments use semi-structured $6{:}8$ sparsity.
Concretely, weights are partitioned into blocks of $M{=}8$ elements and we keep $N{=}6$ weights per block according to a magnitude-based rule, setting the remaining positions to zero.
We refer to our method as \textbf{Sparse-BitNet}, which combines this $6{:}8$ constraint with ternary BitNet-style weight quantization during training.

\noindent\textbf{Training data and objective.}
We train all model sizes on RefineWeb~\citep{penedo2023refinedweb} data for approximately \textbf{50B tokens} per model.
The training objective is standard causal language modeling (next-token prediction).
All reported perplexities are computed on a held-out validation split of the same data distribution.

Across BF16 and BitNet variants, we keep the architecture, data mixture/token budget, optimizer, and learning-rate schedule the same; the key difference is the use of ternary weight/activation quantization operators in BitNet-style training. More training details can be found in Appendix~\ref{app:exp_details}.

\noindent\textbf{Evaluation settings.}
We evaluate model quality using both validation perplexity and downstream task performance.
Specifically, we report accuracy on five widely used benchmarks: HellaSwag~\citep{zellers2019hellaswag}, ARC-E~\citep{clark2018think}, PIQA~\citep{bisk2020piqa}, BoolQ~\citep{clark2019boolq}, and COPA~\citep{roemmele2011choice}.
For efficiency evaluation, we benchmark training and inference throughput on NVIDIA A100 and B200 GPUs.

\subsection{Main Results}
\label{sec:main_results}

\begin{table}[t]
\centering
\caption{\textbf{Perplexity (PPL) degradation analysis.} While BitNet has a higher baseline PPL due to quantization, its \textbf{increase in PPL} after sparsification (values in parentheses) is significantly smaller than that of BF16, highlighting its robustness.}
\label{tab:ppl_singlecol_3sizes}

\small
\setlength{\tabcolsep}{5pt}

\begin{tabular}{l ccc}
\toprule
\multirow{2}{*}{\textbf{Method}} & \multicolumn{3}{c}{\textbf{PPL} $\downarrow$} \\
\cmidrule(l){2-4}
 & \textbf{0.5B} & \textbf{1.5B} & \textbf{3B} \\
\midrule

Dense BF16 & 21.91 & 18.10 & 16.03 \\
Sparse BF16 (6:8) & 23.11 \showgap{1.20}{softred} & 18.70 \showgap{0.60}{softred} & 16.48 \showgap{0.45}{softred} \\
\addlinespace[0.6em]

Dense BitNet & 25.99 & 20.11 & 17.70 \\
Sparse BitNet (6:8) & \textbf{26.31} \textbf{\showgap{0.32}{softblue}} & \textbf{20.35} \textbf{\showgap{0.24}{softblue}} & \textbf{17.87} \textbf{\showgap{0.17}{softblue}} \\

\bottomrule
\end{tabular}
\end{table}

\noindent\textbf{Comparison with baselines.}
We compare \textbf{Sparse-BitNet} against the following baselines across all model sizes:
(i) \textbf{Dense BF16} pretraining without sparsity,
(ii) \textbf{Sparse BF16} pretraining under the same $6{:}8$ sparsity constraint,
(iii) \textbf{Dense BitNet} (ternary) pretraining without sparsity, and
(iv) \textbf{Sparse BitNet} (ours) pretraining under $6{:}8$ sparsity.

\noindent\textbf{Sparse-BitNet is consistently more robust to $6{:}8$ sparsification across scales.}
We evaluate $6{:}8$ semi-structured sparsity on Qwen2.5 models at 0.5B/1.5B/3B and measure sparsity-friendliness by the incremental degradation relative to each method’s own dense baseline ($\Delta$, \cref{tab:ppl_singlecol_3sizes,tab:downstream_delta_soft}). While scaling reduces PPL for both dense and sparse models, enforcing $6{:}8$ induces substantially smaller additional PPL increases for Sparse-BitNet than for BF16 at all scales: BF16 rises by +1.20/+0.60/+0.45 (0.5B/1.5B/3B), whereas Sparse-BitNet increases by only +0.32/+0.24/+0.17. The sparsity penalty for BitNet further shrinks with model size and remains consistently below BF16. Downstream zero-shot results on five benchmarks (HellaSwag, ARC-E, PIQA, BoolQ, COPA) show the same trend: the average accuracy drop under $6{:}8$ is 1.15/3.79/0.80 points for BitNet versus 3.02/7.71/3.20 for BF16 at 0.5B/1.5B/3B. Overall, under identical magnitude-based $N{:}M$ pruning and matched training budgets, ternary BitNet exhibits higher robustness to deployable $6{:}8$ sparsity, i.e., smaller degradation relative to its own dense counterpart (not necessarily higher absolute quality than dense BF16).

\noindent\textbf{BitNet exhibits delayed collapse under increasing semi-structured sparsity.}
\label{sec:main_results_sparsity_sweep}
While our main experiments focus on $6{:}8$ sparse training, we further stress-test robustness on \textsc{Qwen2.5-0.5B} by sweeping a family of $N{:}8$ semi-structured patterns from dense $8{:}8$ to aggressive $2{:}8$.
For each pattern, we train a model \emph{from scratch} under the target $N{:}8$ constraint and report validation perplexity (PPL).
To enable a fair comparison across formats with different absolute PPL scales, we report the \emph{normalized} PPL increase relative to each method's own dense counterpart:
$\mathrm{NormPPL}(N{:}8)=\mathrm{PPL}(N{:}8)/\mathrm{PPL}(8{:}8)$.
Raw PPL values for all patterns are provided in Appendix Table~\ref{tab:raw_ppl_sweep_05b} (also included in our supplementary directory for convenience).

\cref{fig:ppl_sparsity_trend} shows that BF16 deteriorates rapidly as sparsity becomes more aggressive.
Notably, at the hardware-relevant $2{:}4$ setting, the BF16 baseline incurs a $+18.8\%$ normalized PPL increase, exceeding a $10\%$ degradation threshold, whereas BitNet remains stable with only a $+5.7\%$ increase.
Using the $10\%$ threshold as a practical indicator of ``collapse'', BF16 crosses the threshold at $4{:}8$ while BitNet stays below it and only crosses at $3{:}8$.
These results indicate that ternary BitNet models can sustain substantially stronger semi-structured sparsity before quality collapses, yielding a wider feasible sparsity range for deployment.


\begin{wraptable}{r}{0.48\textwidth}
    \vspace{-3em}
    \centering
    \ICMLTableCaption{End-to-End Performance: Qwen2.5-3B (Dense vs. 6:8 Sparse). Throughput is in thousands (k) of tokens/s. \textbf{M} = SeqLen (Prefill) / BatchSize (Decode).}
    \label{tab:llama3_performance}
    
    \renewcommand{\arraystretch}{1.02}
    \setlength{\tabcolsep}{3pt}
    \scriptsize
    
    \begin{tabular}{crrr}
        \toprule
        \multirow{2}{*}{\textbf{}} & \multicolumn{2}{c}{\textbf{Throughput} (k tok/s)} & \multirow{2}{*}{\textbf{Speedup}} \\
        \cmidrule(lr){2-3} 
        \textbf{(M)} & \textbf{Dense} & \textbf{Sparse} & \\
        \midrule
        
        \multicolumn{4}{l}{\cellcolor{gray!10}\textit{\textbf{Prefill (A100)}}} \\ 
        512    & 9.7k  & 10.6k & 1.09$\times$ \\
        1,024  & 20.3k & 21.3k & 1.05$\times$ \\
        2,048  & 37.7k & 42.5k & 1.13$\times$ \\
        4,096  & 40.9k & 52.2k & 1.28$\times$ \\
        8,192  & 42.1k & 53.6k & 1.27$\times$ \\
        16,384 & 42.7k & 55.1k & 1.29$\times$ \\
        32,768 & 42.8k & 55.4k & 1.29$\times$ \\
        65,536 & 42.7k & 55.5k & \textbf{1.30$\times$} \\
        
        \midrule
        
        \multicolumn{4}{l}{\cellcolor{gray!10}\textit{\textbf{Decode (B200)}}} \\ 
        64     & 11.1k & 12.2k & 1.09$\times$ \\
        128    & 17.2k & 20.4k & \textbf{1.18$\times$} \\
        256    & 25.9k & 29.1k & 1.12$\times$ \\
        512    & 30.4k & 34.4k & 1.13$\times$ \\
        \bottomrule
    \end{tabular}
    \vspace{-4em}
\end{wraptable}

\noindent\textbf{Inference speed up.}
We implemented an in-house 6:8 sparse kernel for acceleration and report performance metrics for the prefill phase on an NVIDIA A100 GPU and the decoding phase on a B200 GPU.
\cref{tab:llama3_performance} details the raw throughput (tokens/s) and achieved speedup of the 6:8 sparse Qwen2.5-3B model across various input configurations (denoted as $M$, representing sequence length and batch size).

\subsection{Ablation Studies}
\label{sec:ablation}

\noindent\textbf{Optimization design for dynamic $6{:}8$ masks.}
\label{sec:gradient}

\begin{figure}[t]
    \centering
    \begin{subfigure}[t]{0.49\textwidth}
        \centering
        \includegraphics[width=\linewidth]{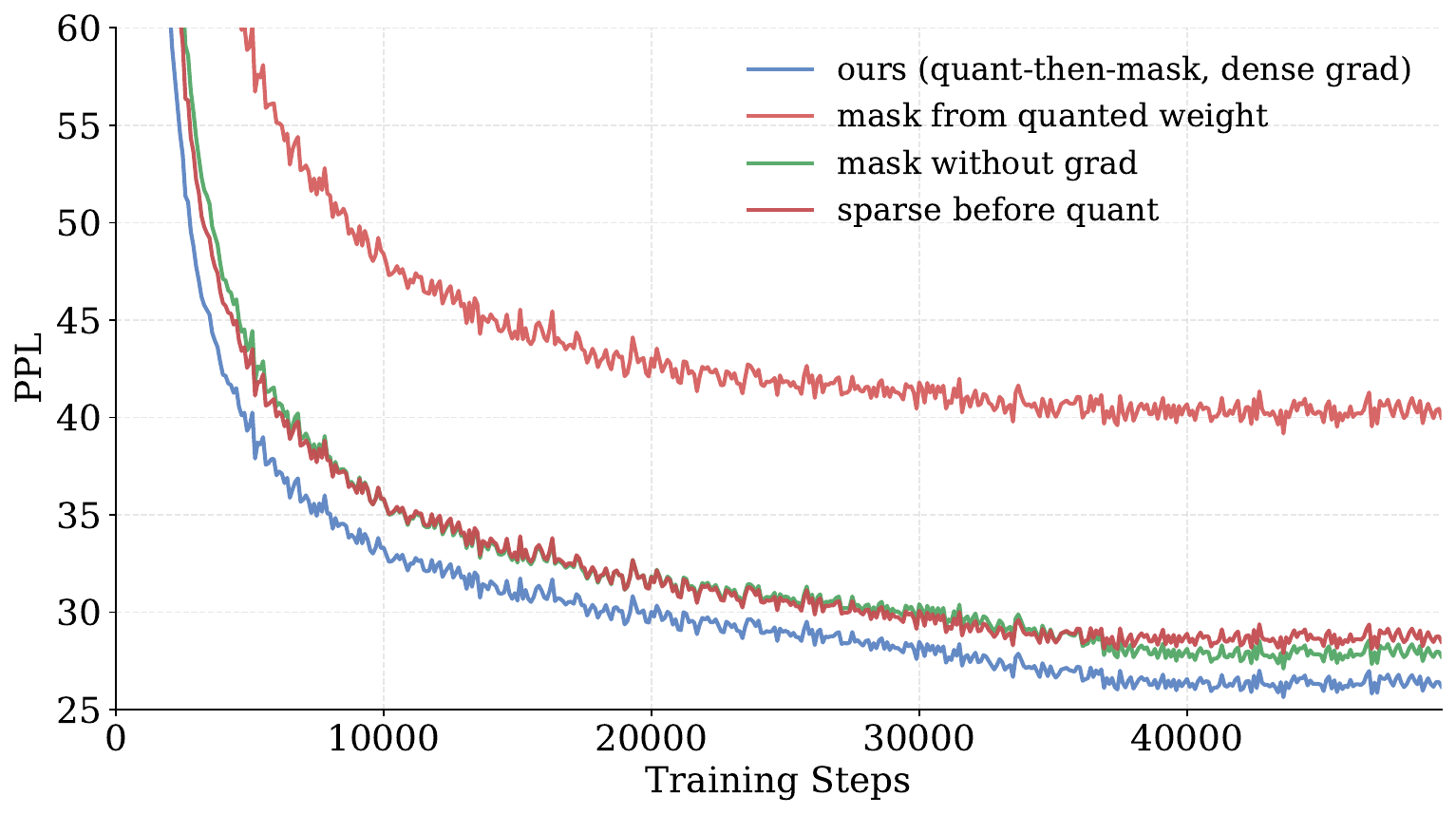}
        \caption{Validation perplexity (lower is better).}
        \label{fig:ablation_ppl}
    \end{subfigure}
    \hfill
    \begin{subfigure}[t]{0.49\textwidth}
        \centering
        \includegraphics[width=\linewidth]{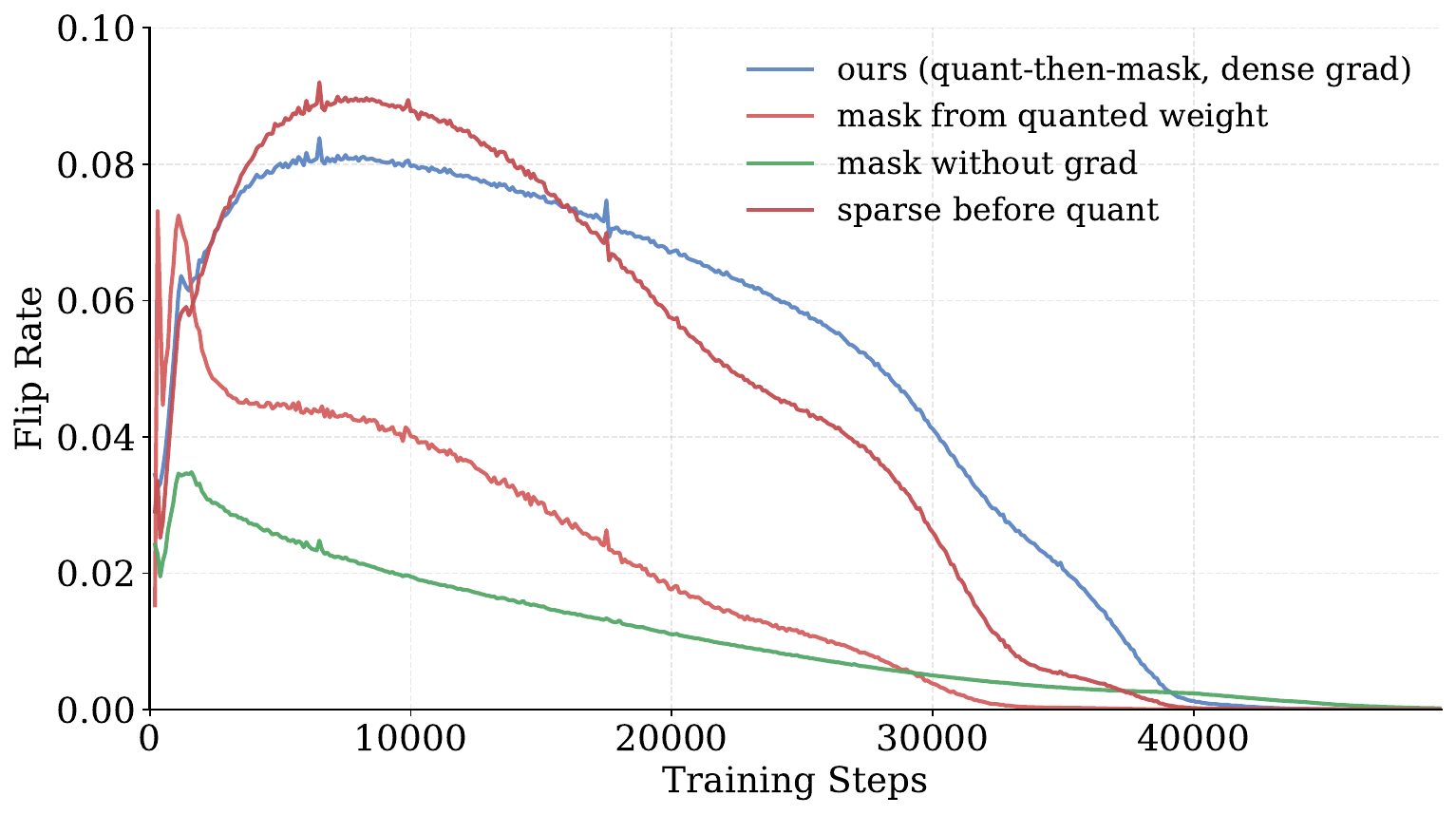}
        \caption{Mask flip rate $r_t$ (Eq.~\ref{flip_rate}).}
        \label{fig:ablation_flip_rate}
    \end{subfigure}

    \caption{\textbf{Ablation on training design choices for dynamic $6{:}8$ sparsity (Qwen2.5-0.5B).}(a) Validation perplexity curves under different training design choices.
(b) Corresponding mask flip rate $r_t$ (Eq.~\ref{flip_rate}), reflecting the stability of sparsity pattern evolution during training.}
    \label{fig:ablation_ppl_flip}
\end{figure}

Sparse-BitNet jointly optimizes ternary weights and semi-structured $6{:}8$ connectivity.
At each step, we generate a binary mask $m(\cdot)$ from the current full-precision master weights and apply it to enforce the $6{:}8$ constraint in the forward pass.
Because both ternary quantization and dynamic top-$N$ selection are non-differentiable, a robust training recipe hinges on three coupled design choices:
(i) whether masked weights can still receive gradient updates,
(ii) whether masks are constructed from continuous master weights or from discrete ternary weights, and
(iii) whether masking is applied before or after ternary quantization.

We compare the following four variants under identical training settings on \textsc{Qwen2.5-0.5B} with $6{:}8$ sparsity (see \cref{fig:ablation_ppl,fig:ablation_flip_rate}):

\begin{itemize}
    \item \textbf{Baseline (ours): quant-then-mask + dense gradient flow.}
    We compute the mask from the master weights $w$ (continuous ranking), apply ternary quantization $Q_t(\cdot)$, and enforce sparsity on the discrete weights:
    $W_{\text{eff}} = m(w)\odot Q_t(w)$.
    During backprop, we use straight-through estimators so that \emph{all} master weights (including currently masked ones) receive gradient updates.
    
    \item \textbf{Mask without grad (stop-grad on masked weights).}
    Same forward computation as the baseline, but we block gradients on masked entries (i.e., multiply gradients by the current mask), so masked weights are not updated.
    
    \item \textbf{Mask from quantized weight (quantized-mask selection).}
    We first quantize the master weights and then construct the $6{:}8$ mask from the ternary weights, i.e., use $m(Q_t(w))$ for top-$N$ selection.
    The forward pass becomes $W_{\text{eff}} = m(Q_t(w))\odot Q_t(w)$.
    
    \item \textbf{Sparse before quant (mask-then-quant).}
    We first apply the $6{:}8$ mask to the master weights and then quantize only the masked weights:
    $W_{\text{eff}} = Q_t(m(w)\odot w)$.
\end{itemize}

\noindent\textbf{Perplexity and convergence.}
\cref{fig:ablation_ppl} shows that our baseline achieves the best convergence and the lowest final perplexity.
Blocking gradients on masked weights consistently hurts optimization, leading to a higher PPL.
More strikingly, constructing masks from \emph{quantized} ternary weights severely destabilizes training and fails to reach a competitive perplexity, plateauing at a much worse level.
Finally, applying sparsity before quantization (mask-then-quant) is also inferior to the baseline, indicating that the order in which quantization and masking are composed matters for stable optimization.

\noindent\textbf{Mask dynamics via flip rate.}
To understand the optimization dynamics, we monitor the \emph{mask flip rate} following prior work~\cite{hu2024accelerating}.
Let $w_t$ be a $D$-dimensional weight vector at step $t$ and $m(w_t)\in\{0,1\}^D$ its corresponding $6{:}8$ mask.
We define the flip rate as
\begin{equation}
r_t \;=\; \frac{\|m(w_t) - m(w_{t-1})\|_1}{D} \in [0,1].
\label{flip_rate}
\end{equation}
Intuitively, $r_t$ measures how frequently the sparse connectivity pattern changes.
A healthy sparse training process typically exhibits an exploration-to-convergence behavior: non-trivial flips early on (exploring mask configurations) followed by a gradual decay as masks stabilize.

As shown in \cref{fig:ablation_flip_rate}, our baseline follows this desirable trajectory, with substantial exploration early and smooth stabilization later.
In contrast, \textbf{mask without grad} shows a much lower flip rate throughout training, suggesting premature mask freezing: once weights are masked, they cannot be updated to re-enter the top-$N$ set, which limits exploration and degrades the final solution.
The \textbf{mask from quantized weight} variant exhibits persistently large and noisy flip rates early on, consistent with unstable mask selection.
This behavior is expected because ternary quantization creates many ties (e.g., many weights at 0 or $\pm 1$), making top-$N$ selection sensitive to small perturbations and tie-breaking, which in turn prevents stable convergence.
Finally, \textbf{sparse before quant} reduces the flip rate relative to the baseline and converges to a worse perplexity, indicating that quantizing after masking can undesirably couple quantization noise/scale estimation with the current sparse subset, hindering effective exploration.

\noindent\textbf{Takeaway.}
Together, these ablations justify our final training design: (i) allow gradients to update masked master weights so that the model can continuously explore and revise sparse connectivity, (ii) construct masks from continuous master weights rather than ternary weights to avoid tie-driven instability, and (iii) enforce sparsity via a quant-then-mask forward pass to produce well-defined $6{:}8$ sparse discrete weights for inference while retaining stable optimization behavior during training.

\noindent\textbf{Dense-to-sparse training schedule.}
We next study whether the \emph{training trajectory} affects the final quality under semi-structured sparsity.
Concretely, we adopt a two-stage schedule where we first train densely and then switch to $6{:}8$ sparse training for the remaining steps.
Let $\rho \in \{0,25,50,75,100\}$ denote the fraction (\%) of the total training budget spent in the sparse phase, where $\rho{=}0$ corresponds to dense-only training and $\rho{=}100$ corresponds to sparse-from-scratch.

\begin{figure}[t]
    \centering
    \begin{subfigure}[t]{0.48\linewidth}
        \centering
        \includegraphics[width=\linewidth]{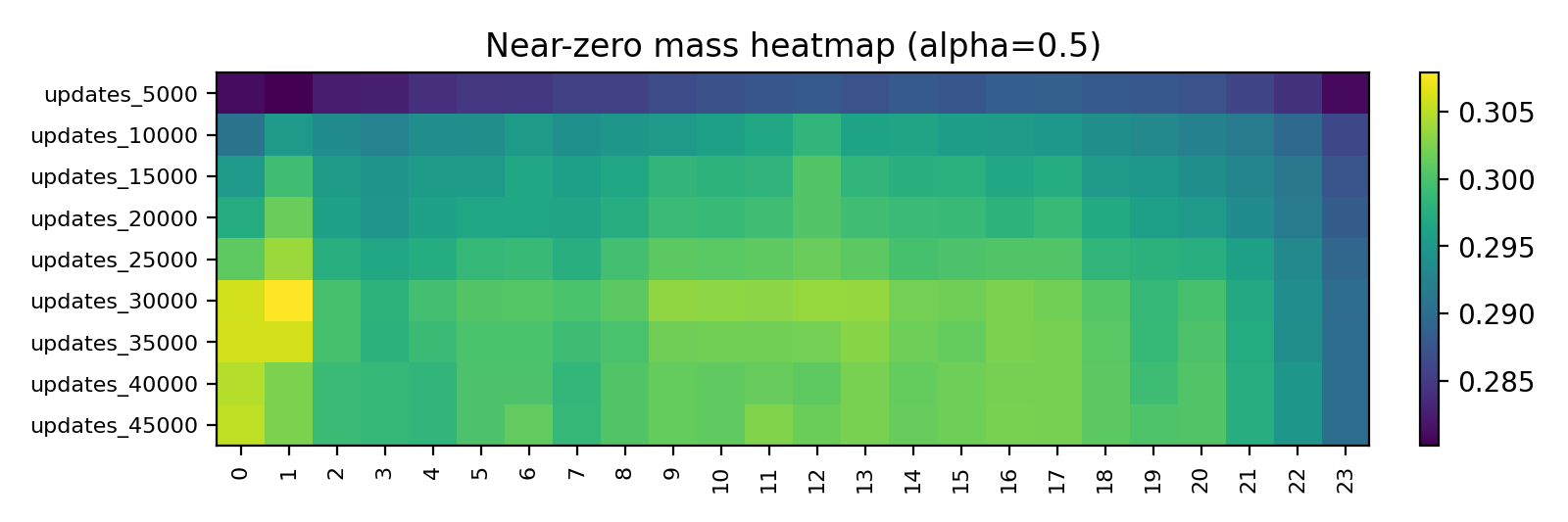}
        \caption{BF16 dense: near-zero mass (master weights).}
        \label{fig:bf16_heatmap}
    \end{subfigure}
    \hfill
    \begin{subfigure}[t]{0.48\linewidth}
        \centering
        \includegraphics[width=\linewidth]{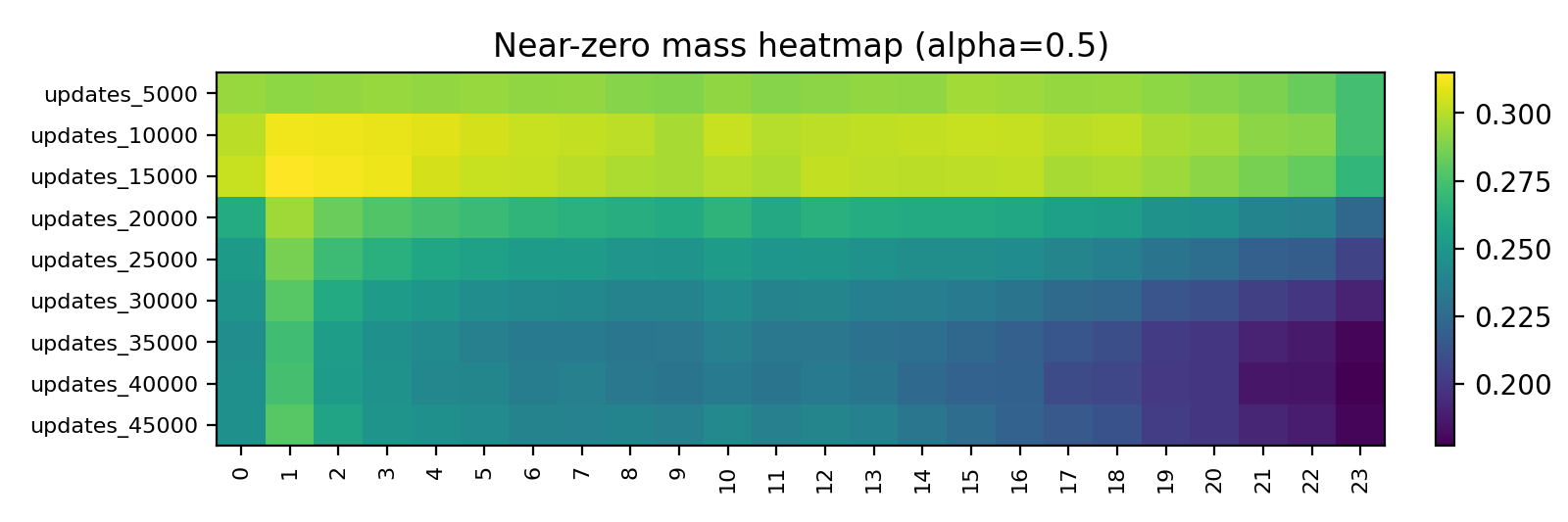}
        \caption{BitNet dense: near-zero mass (master weights).}
        \label{fig:bitnet_heatmap}
    \end{subfigure}
    
    \caption{\textbf{Polarization trend under ternary QAT.} 
    (a) BF16 maintains a concentration around zero. 
    (b) BitNet shows decreasing near-zero mass, indicating strong polarization over time.}
    \label{fig:heatmap_comparison}
\end{figure}

\begin{figure}[t]
    \centering
    \includegraphics[width=\linewidth]{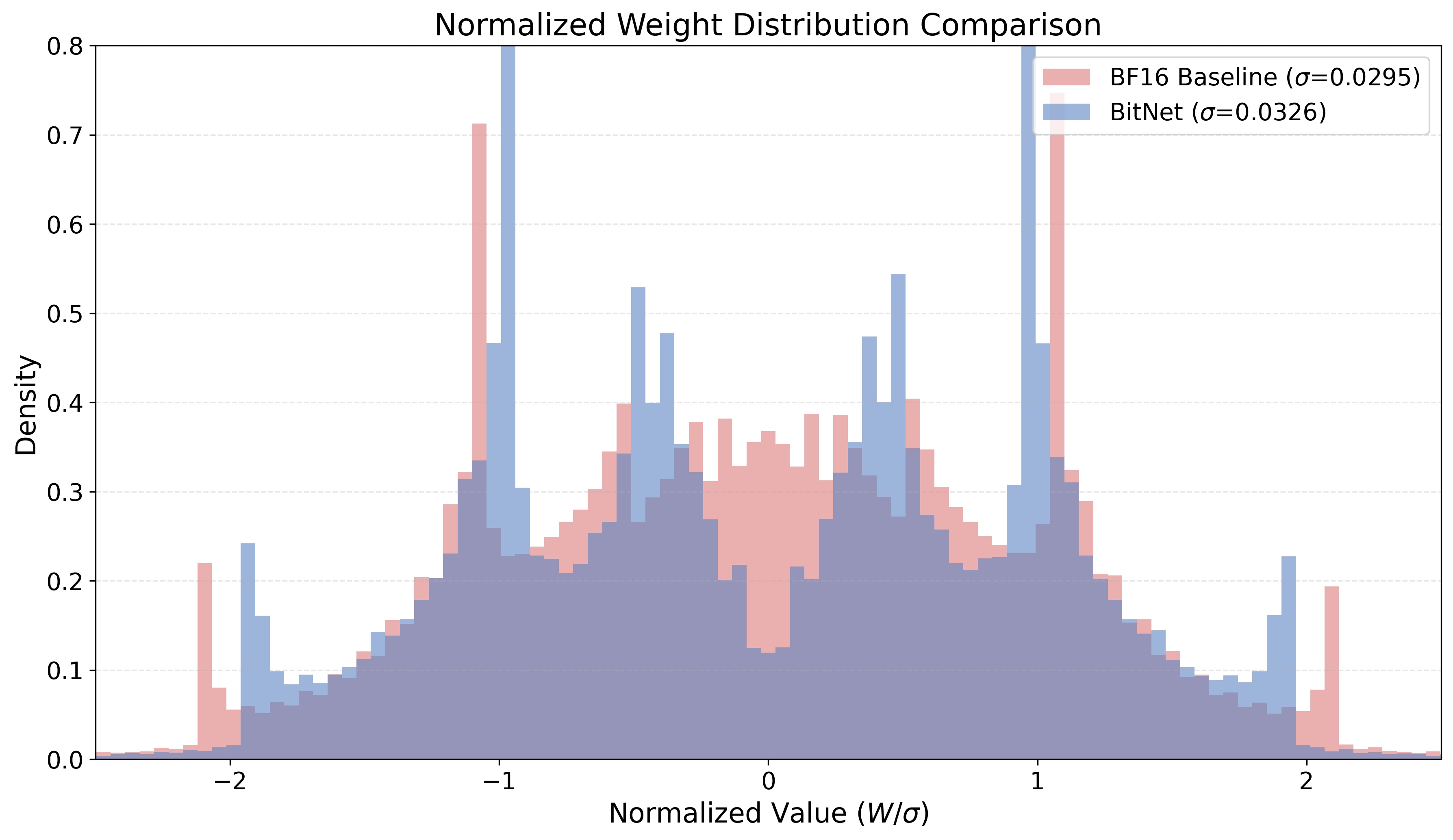}
    
    \caption{\textbf{Weight Distribution.} 
    Global histogram of linear-layer master weights at the final checkpoint. 
    Unlike the unimodal distribution of BF16, BitNet displays a structured, multi-modal magnitude landscape, confirming the intrinsic sparsity property.}
    \label{fig:weight_histogram}
\end{figure}

\begin{figure*}[t]
\centering
\begin{subfigure}{0.49\linewidth}
  \centering
  \includegraphics[width=\linewidth]{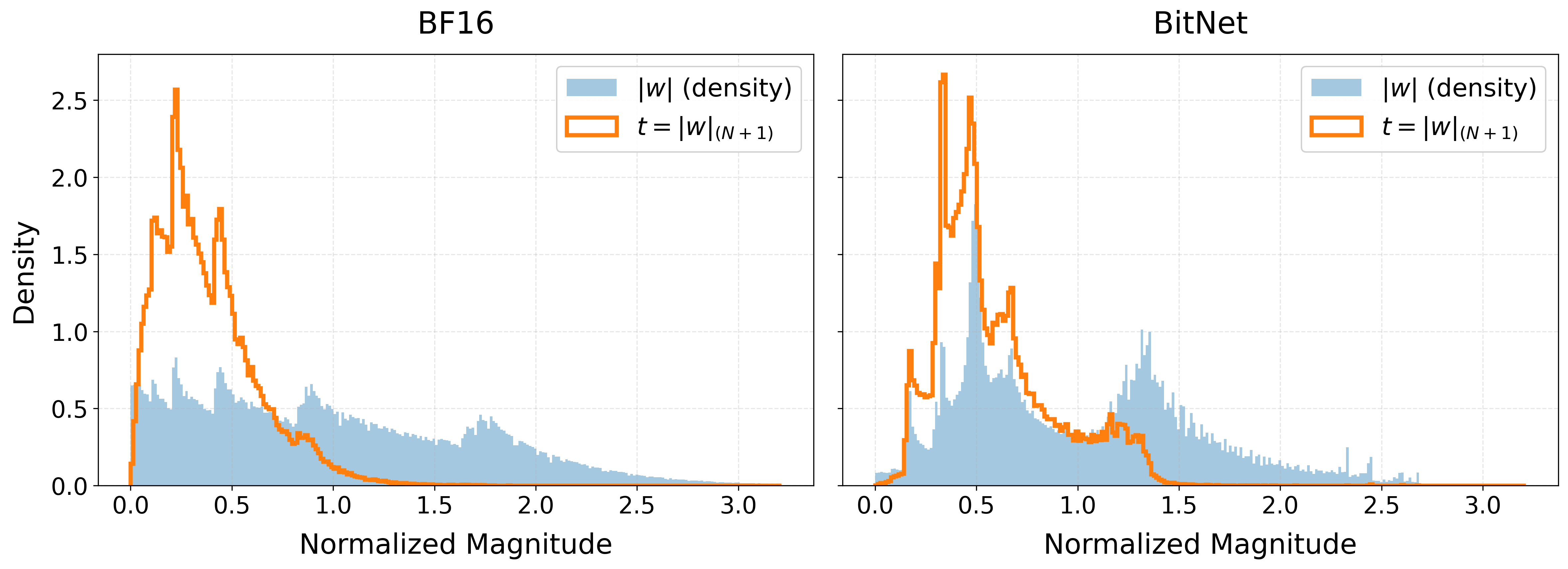}
  \caption{Mid layers.}
\end{subfigure}
\begin{subfigure}{0.49\linewidth}
  \centering
  \includegraphics[width=\linewidth]{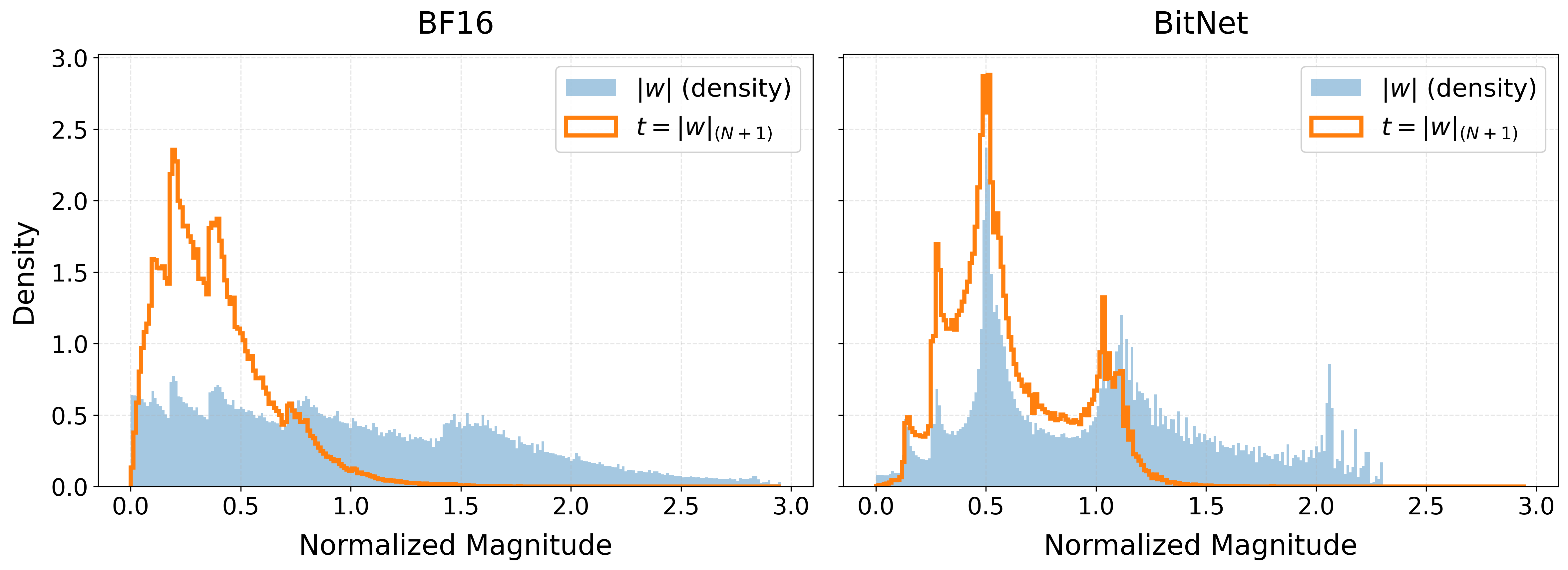}
  \caption{Late layers.}
\end{subfigure}
\caption{\textbf{Overlay of $|w|$ density and per-block threshold $t$ (dense training).}
Blue: density of normalized magnitudes $|w|$; Orange: density of thresholds $t=|w|_{(N+1)}$ for $6{:}8$ selection.
BitNet shows a more pronounced higher-magnitude population in mid/late layers, while thresholds concentrate mainly in the low-to-mid regime, suggesting that $N{:}M$ selection primarily operates within lower-magnitude candidates.}
\label{fig:analysis_threshold_overlay}
\end{figure*}

As shown in \cref{tab:ablation_schedule_ppl}, switching to sparsity late significantly worsens convergence.
In particular, when only the last $25\%$ or $50\%$ of steps are trained sparsely, validation PPL degrades to $27.48$ and $27.39$, respectively, compared to $26.31$ for sparse-from-scratch.
Increasing the sparse budget improves the outcome: using $75\%$ sparse steps reduces PPL to $26.71$, but remains worse than training sparsely throughout.
Overall, these results indicate that effective $6{:}8$ training requires substantial sparse adaptation budget; delaying the switch to sparsity yields a persistent quality gap under the same total training budget.

\begin{table}[t]
\ICMLTableCaption{\textbf{Effect of dense-to-sparse schedule on validation PPL} (6:8, \textsc{Qwen2.5-0.5B}). 
$\rho$ denotes the fraction of total training steps using $6{:}8$ sparse training after an initial dense phase.}
\label{tab:ablation_schedule_ppl}

\centering
\small
\setlength{\tabcolsep}{6pt}
\begin{tabular}{lc}
\toprule
Sparse ratio $\rho$ (\%) & Val PPL$\downarrow$ \\
\midrule

25                   & 27.48 \\
50                   & 27.39 \\
75                   & 26.71 \\
100 (sparse-from-scratch) & 26.31 \\
dense training     & 25.99 \\
\bottomrule
\end{tabular}
\end{table}

\noindent\textbf{Mask-from-master vs.\ mask-from-quantized.}
A second key design choice in Sparse-BitNet is \emph{which} weights are used to construct the $6{:}8$ pruning mask.
Our default implementation computes the mask from the full-precision \emph{master} weights (BF16), and then applies ternary quantization on the masked weights in the forward pass.
An alternative is to compute the mask directly from the \emph{quantized} ternary weights.
While seemingly natural, this choice can be problematic because ternary weights introduce many ties in magnitude (e.g., $|w|\in\{0,1\}$), making top-$|w|$ selection within each $8$-element block ill-conditioned and potentially unstable.

We evaluate these two options on \textsc{Qwen2.5-0.5B} with $6{:}8$ sparse training.
Using master weights to generate the mask yields a validation PPL of $26.31$, whereas generating the mask from quantized weights causes a drastic degradation to $32.23$.
This large gap confirms that mask selection should be performed in the continuous master-weight space, where magnitudes provide a reliable ranking signal; in contrast, quantized-space masking suffers from severe tie effects and leads to substantially worse optimization and final quality.

\subsection{Analysis}
\label{sec:analysis}

\paragraph{Ternary QAT induces polarization rather than unimodal shrinkage.}
To better understand why Sparse-BitNet exhibits superior robustness under $6{:}8$ sparsity, we analyze the dense training dynamics of latent weights prior to any explicit pruning.
We track the \emph{near-zero mass}, defined as $\mathbb{P}(|w|/\mathrm{mean}(|w|) < 0.5)$.
As shown in \cref{fig:heatmap_comparison}, the BF16 baseline exhibits a sustained concentration near zero, forming a unimodal distribution where "important" and "redundant" weights are structurally entangled.
In contrast, BitNet shows a clear polarization trend: the near-zero mass decreases significantly over time, indicating that latent weights actively migrate away from the ambiguous region toward decisive magnitudes.
This results in a structured, tri-modal histogram (\cref{fig:weight_histogram}) featuring distinct "active" clusters separated from the "dead" zone.
This suggests that BitNet's dense optimization naturally acts as a soft-selector, pre-sorting weights into a topology compatible with sparsity.

\paragraph{Per-block thresholds are decoupled from active weights in BitNet.}
Sparsity is locally determined by the per-block threshold $t = |w|_{(N+1)}$.
A key question is whether this local threshold "cuts into" the signal or safely removes noise.
\cref{fig:analysis_threshold_overlay} overlays the distribution of normalized weights $|w|$ (blue) with the distribution of pruning thresholds $t$ (orange).

For BF16 (Left), the threshold distribution closely shadows the weight distribution. This implies a strong coupling: the pruning boundary frequently intersects with the main body of the weight population, making the removal of important information inevitable.
Conversely, BitNet (Right, especially Late Layers) exhibits a phenomenon we term magnitude stratification.
The weight distribution (blue) develops a secondary "active" mode at higher magnitudes (e.g., $x \in [0.5, 1.5]$). Crucially, the threshold distribution (orange) remains concentrated in the lower regime ($x < 0.5$) and drops off sharply before reaching the active mode.
This indicates a structural decoupling: the $N{:}M$ selection boundary operates almost exclusively within the noise/redundant parameter space, leaving the population of high-magnitude weights largely intact.
This mechanism explains why Sparse-BitNet sustains structural integrity even under aggressive pruning constraints.

\section{Related Works}




\subsection{Quantization and BitNet Architectures}

Quantization methods~\citep{an2024fluctuation, chee2023quip, lee2023owq, liu2024llm, du2024bitdistiller} optimize efficiency by compressing weights (e.g., from 16-bit to 8-bit) via either training from scratch or converting pre-trained models. We collectively refer to extreme binary or ternary quantization as BitNet. While standard BitNet~\cite{wang2023bitnet} constrains weights to 1-bit, BitNet b1.58~\citep{ma2024era} employs a ternary $\{-1, 0, 1\}$ scheme. By incorporating zero, this approach significantly enhances capacity, achieving performance comparable to full-precision models~\citep{wang2024bitnet,ma2025bitnetb158,wang2025bitnet} and effectively bridging the gap between extreme quantization and standard architectures.

\subsection{Model Pruning}

Model pruning is broadly categorized into unstructured and structured approaches. Unstructured pruning, pioneered by \cite{han2015learning}, targets individual weights with low magnitudes. While recent adaptations for LLMs~\cite{jaszczur2021sparse, sun2023simple,dong2024pruner,zhang2024plug} achieve high sparsity with minimal performance loss, they often lack direct hardware acceleration. To bridge this gap, N:M sparsity~\cite{mishra2021accelerating} introduces fine-grained structural constraints to enable hardware-friendly acceleration.

Conversely, structured pruning explicitly targets hardware efficiency by removing coherent architectural components, such as attention heads or rows/columns~\cite{ma2023llm,xia2023sheared,an2024fluctuation,ashkboos2024slicegpt}. Although this guarantees inference speedup, it typically incurs significant accuracy degradation, necessitating expensive retraining.

Crucially, both paradigms generally operate as post-hoc optimizations on pre-trained dense models. Distinct from these approaches, our method introduces sparsity dynamically during training. This paradigm shift enables us to leverage sparsity not only for inference deployment but also to enhance training efficiency significantly.

\section{Conclusion}
\label{sec:conclusion}

We presented \textbf{Sparse-BitNet}, a unified framework establishing that $1.58$-bit ternary LLMs are intrinsically more robust to semi-structured sparsity than their BF16 counterparts. 
By integrating dynamic $6{:}8$ masking with ternary quantization, our method significantly reduces performance degradation and delays model collapse across varying scales. 
Our ablation studies highlight that computing masks from dense master weights and maintaining gradient flow through masked regions are critical for effective optimization. 
Furthermore, our custom $6{:}8$ kernels demonstrate practical inference speedups of up to $1.30\times$, confirming that combining extreme quantization with structured pruning offers a viable Pareto frontier for efficient LLM deployment. 

\appendix
\onecolumn

\section{Experiment Setup}
\label{app:exp_details}



\subsection{Training hyperparameters}
\label{app:hyperparams}
Table~\ref{tab:hyperparams} summarizes the full training hyperparameters used for all runs.

\begin{table}[h]
\centering
\small
\caption{Training hyperparameters for all experiments.}
\begin{tabular}{l c}
\toprule
Hyperparameter & Value \\
\midrule
Optimizer & AdamW \\
$\beta_1, \beta_2, \epsilon$ & [$0.9$, $0.95$, $1e-5$] \\
Learning rate & 1e-5 \\
Schedule & cosine \\
Warmup ratio & 0.5 \\
Weight decay & 0.1 \\
Micro-batch size & 16 \\
Gradient accumulation & 4 \\
Sequence length & 2048 \\
Gradient clipping & 1.0 \\
Precision & BF16 \\

\bottomrule
\end{tabular}
\label{tab:hyperparams}
\end{table}

\subsection{Baselines and fairness controls}
\label{app:fairness}
\paragraph{Matched training budget.}
All variants are trained with identical tokens, data mixture, and optimizer settings.
We only change the components under study (sparsity/quantization), and keep architecture, initialization, and evaluation protocol fixed.

\paragraph{Baseline implementations.}
Structured sparse baselines use custom kernel with the same $6{:}8$ pattern and weight layout for main results.

\section{Pseudo torch-style implementation of Sparse-BitLinear.}
\label{app:Sparse-BitLinear}
Algorithm~\ref{alg: sparse_bitlinear} presents the core logic of our proposed Sparse-BitLinear architecture. It highlights the sequence of pre-quantization masking and the subsequent ternary scaling process. The implementation of the WeightQuantMasked autograd function further clarifies how the straight-through estimator (STE) is applied to maintain dense gradient flow during backpropagation, a key factor in mitigating sparsity-induced degradation.


\begin{algorithm*}[h]
\scriptsize
\caption{Pseudo-code for Sparse-BitLinear with Explicit Dual-STE Backward.}
\label{alg: sparse_bitlinear}
\definecolor{codeblue}{rgb}{0.25,0.5,0.8}
\definecolor{codegreen}{rgb}{0,0.6,0}
\definecolor{codekw}{rgb}{0.85, 0.18, 0.50}
\lstset{
  backgroundcolor=\color{white},
  basicstyle=\fontsize{7.5pt}{7.5pt}\ttfamily\selectfont,
  columns=fullflexible,
  breaklines=true,
  captionpos=b,
  commentstyle=\fontsize{7.5pt}{7.5pt}\color{codeblue},
  keywordstyle=\fontsize{7.5pt}{7.5pt}\color{codekw},
  escapechar={|}, 
}
\begin{lstlisting}[language=python]
# Custom Autograd Function to define explicit Forward/Backward behavior
class WeightQuantMasked(torch.autograd.Function):
    @staticmethod
    def forward(ctx, w, N, M):
        # 1. Dynamic N:M Masking (Pre-Quant)
        # Mask is computed on dense weights |w| to preserve ranking
        mask = compute_nm_mask(w, N, M) 
        

        # 2. Ternary Quantization (1.58-bit)
        # Scale -> Round -> Clamp -> Rescale
        scale = w.abs().mean().clamp(min=1e-5)
        w_q = (w_masked / scale).round().clamp(-1, 1) * scale
        w_masked = w_q * mask
        return w_masked

    @staticmethod
    def backward(ctx, g_out):
        # Dual-STE Implementation:
        # Gradient flows straight through to dense weights, ignoring
        # both the N:M mask (zeros) and quantization rounding.
        # g_w = g_out * 1.0
        return g_out, None, None

class SparseBitLinear(nn.Linear):
    def __init__(self, in_features, out_features, N=2, M=4):
        super().__init__(in_features, out_features)
        self.N, self.M = N, M

    def forward(self, x):
        # 1. Activation Quantization (8-bit AbsMax)
        s_x = 127.0 / x.abs().max(dim=-1, keepdim=True).values.clamp(min=1e-5)
        x_q = (x * s_x).round().clamp(-128, 127) / s_x

        # 2. Weight Quantization & Masking (with STE)
        # Apply the custom autograd function
        w_q = WeightQuantMasked.apply(self.weight, self.N, self.M)

        # 3. Linear Projection
        return F.linear(x_q, w_q, self.bias)
\end{lstlisting}
\end{algorithm*}

\section{Raw PPL for sparsity sweep}
The complete validation perplexity (PPL) statistics for the N:8 sparsity training are summarized in \cref{tab:raw_ppl_sweep_05b}.
\begin{table}[h]
\ICMLTableCaption{\textbf{Raw validation PPL for the $N{:}8$ sparsity sweep on \textsc{Qwen2.5-0.5B}.}}
\label{tab:raw_ppl_sweep_05b}

\centering
\small
\setlength{\tabcolsep}{6pt}
\begin{tabular}{lccccccc}
\toprule
Method & 8:8 & 7:8 & 6:8 & 5:8 & 2:4 & 3:8 & 2:8 \\
\midrule
BitNet & 25.99& 26.12& 26.31& 26.71& 27.48& 29.80& 33.12\\
BF16 & 21.91& 22.27& 23.11& 23.42& 26.03& 28.66& 31.70\\
\bottomrule
\end{tabular}
\end{table}
\label{sec:appendix}

\bibliographystyle{alpha} 
\bibliography{main}

\end{document}